\pdfoutput=1

\documentclass{article}
\usepackage[preprint,nonatbib]{neurips_2020}

\usepackage[utf8]{inputenc} 
\usepackage[T1]{fontenc}    
\usepackage{hyperref}       
\usepackage{url}            
\usepackage{booktabs}       
\usepackage{amsfonts}       
\usepackage{nicefrac}       
\usepackage{microtype}      

\usepackage{graphicx}
\usepackage{amsmath,amssymb} 
\usepackage{color}
\usepackage{hyperref}
\usepackage[capitalise]{cleveref}
\usepackage{wrapfig}
\usepackage{bm}
\usepackage{bbm}
\usepackage{algpseudocode}
\usepackage{algorithm}
\usepackage{subfig}
\algnewcommand\algorithmicinput{\textbf{Input:}}
\algnewcommand\Input{\item[\algorithmicinput]}

\newcommand{\mx}{\bm{X}}
\newcommand{\my}{\bm{Y}}
\newcommand{\mz}{\bm{Z}}
\newcommand{\vx}{\bm{x}}
\newcommand{\vz}{\bm{z}}
\newcommand{\vsig}{\bm{\Sigma}}
\newcommand{\vmu}{\bm{\mu}}
\newcommand{\vth}{\bm{\theta}}
\newcommand{\vphi}{\bm{\Phi}}
\newcommand{\vw}{\bm{w}}
\newcommand{\norm}[2]{\mathcal{N}(#1#2)}
\newcommand{\normst}{\norm{\bm{0}, \bm{I}}}
\newcommand{\cl}{\mathcal{L}}

\newcommand{\cd}{\mathcal{D}}

\newcommand{\bbe}{\mathbb{E}}
\newcommand{\bbr}{\mathbb{R}}

\newcommand{\px}{p\left( \vx \right)}
\newcommand{\pz}{p\left( \vz \right)}
\newcommand{\ppxz}{p_{\vth}\left( \vx|\vz \right)}
\newcommand{\pxz}{p\left( \vx|\vz \right)}
\newcommand{\ppyz}{p_{\vw}\left( y|\vz \right)}
\newcommand{\pyz}{p\left( y|\vz \right)}
\newcommand{\pzy}{p\left( \vz | y \right)}
\newcommand{\pzxy}{p\left( \vz | \vx, y \right)}
\newcommand{\ppzy}{p_{\vw}\left( \vz | y \right)}
\newcommand{\qqzx}{q_{\vphi}\left( \vz | \vx \right)}
\newcommand{\qzx}{q\left( \vz | \vx \right)}

\newcommand{\qzxy}{q\left( \vz | \vx, y \right)}
\newcommand{\pxay}{p\left( \vx, y \right)}
\newcommand{\pxyz}{p(\vx, y|\vz)}
\newcommand{\pzaxay}{p\left( \vz, \vx, y \right)}
\newcommand{\kl}[1]{D_{KL}\left( #1 \right)}
\newcommand{\lr}[1]{\left(}
\newcommand{\rr}[1]{\right)}
\newcommand{\ls}[1]{\left[}
\newcommand{\rs}[1]{\right]}

\newcommand{\eye}{\textbf{Eye}}
\newcommand{\fac}{\textbf{Facial}}
\newcommand{\efac}{\textbf{EyeFacial}}

\DeclareMathOperator*{\argmax}{arg\,max}
\DeclareMathOperator*{\argmin}{arg\,min}

\begin{document}

\title{IntroVAC: Introspective Variational Classifiers for Learning Interpretable Latent Subspaces} 
\author{Marco Maggipinto \\
  Department of Information Engineering\\
  University of Padova\\
  \href{mailto:marco.maggipinto@dei.unipd.it}{\texttt{marco.maggipinto@dei.unipd.it}} \\
  \And
  Matteo Terzi  \\
  Department of Information Engineering\\
  University of Padova\\
    \href{mailto:terzimat@dei.unipd.it}{\texttt{terzimat@dei.unipd.it}} \\
  \AND
  Gian Antonio Susto \\
  Department of Information Engineering\\
  University of Padova\\
  \href{mailto:gianantonio.susto@unipd.it}{\texttt{gianantonio.susto@unipd.it}} \\
}

\maketitle

\begin{abstract}
Learning \emph{useful}  representations of complex data has been the subject of extensive research for many years. With the diffusion of Deep Neural Networks, Variational Autoencoders have gained lots of attention since they provide an explicit model of the data distribution based on an encoder/decoder architecture which is able to both generate images and  encode them in a low-dimensional subspace. However, the latent space is not easily interpretable and the generation capabilities show some limitations since images typically look blurry and lack details. In this paper, we propose the Introspective Variational Classifier (IntroVAC), a model that learns interpretable latent subspaces by exploiting information from an additional label and provides improved image quality thanks to an adversarial training strategy. 
We show that IntroVAC is able to learn meaningful directions in the latent space enabling fine-grained manipulation of image attributes. We validate our approach on the CelebA dataset.
\end{abstract}

\section{Introduction} \label{intro}
Deep Generative Models have shown amazing performance at modeling complex probability distributions with support on high dimensional manifolds and they are nowadays widespread used for image generation \cite{brock2018large} and representation learning \cite{higgins2016beta}. Variational Autoencoders (VAEs) \cite{kingma2013auto} are appealing models for both tasks since their architecture is based on two Deep Neural Networks (DNNs), one that acts as an encoder and is able to provide low dimensional representations of the input data and one that acts as a decoder allowing to generate data in the input space by starting from their latent representations. Understanding the latent subspace of VAEs has been, since their publication, the subject of interests for many researchers. In particular, in order to get a better understanding of the functioning of such complex models, there has been a lot of focus on learning disentangled representations that are related to visually interpretable aspects (e.g. color, shape)~\cite{burgess2018understanding}. 
However, on one side, how to obtain human-aligned representations without any supervision is still an open challenge and on the other side, understanding and describing representations provided by VAEs is often difficult.
Moreover, the images provided  by such architectures are typically much less realistic than the ones obtained with implicit models such as Generative Adversarial Networks (GANs) \cite{goodfellow2014generative} that, however, cannot be used for representation learning since they only employ a deterministic generator that maps a prior distribution to the data distribution. For these reasons, in this work we propose a different (but in some sense equivalent) task from the one of learning disentangled representations: learning directions in the latent subspace of VAEs that are correlated to a class label assumed as known during training. To solve this task, we equip VAEs with an additional random variable representing the class label in a model called Introspective Variational Classifier (IntroVAC) that has limited increase in complexity compared to standard VAEs. In fact, our model can be trained by maximizing a suitable Evidence Lower Bound (ELBO) plus an additional adversarial loss used to improve image quality. Note that IntroVAC do not require additional discriminators but simply exploits a linear classifier in the feature space to both provide interpretable subspaces and set up an adversarial game between encoder and decoder. 

\subsection{Contributions and related works}
The contributions of the present work can be summarized as follows:
\begin{itemize}
\item We propose IntroVAC to learn interpretable latent subspaces in VAEs exploiting information related to a class label;
\item We propose to learn directions in the latent space that are correlated to class attributes, differently from typical works that aim at learning disentangled representations;
\item We exploit a simple linear classifier to both learn such directions and also to let the encoder and decoder engage in an adversarial game that allows to improve image quality.
\end{itemize}
Learning useful representations with DNNs has been the subject of  research  for many years  \cite{hinton2006reducing,vincent2011connection}, typically encoder/decoder-based architectures have been exploited to map the data in a low-dimensional latent space and reconstruct them. VAEs \cite{kingma2013auto} first provided a generative model with a formal probabilistic derivation that results in an autoencoding architecture; after their publication, a plethora of new works based on \cite{kingma2013auto}  have been developed that exploit such architecture for particular tasks or extend it \cite{sonderby2016ladder,chen2018isolating,kim2018semi}. The closest to our work are Conditional Subspace VAEs (CS-VAE)  \cite{klys2018learning} and Guided VAEs \cite{ding2020guided}. They both rely on a class label to disentangle a predefined subset of latent variables from the others in a way that information on the class label remains only on such subset. The task is still related to disentangled representation learning and the models all suffer from poor image quality as VAEs. Our work differs on many aspects: \begin{itemize}
    \item We don't aim at learning disentangled representations but directions in the latent space related to the class attributes;
    \item Both CS-VAE and Guided-VAE require an additional discriminator while our model only have a linear classifier on top of the latent features that we exploit to provide an interpretable latent space and improved image quality;
    \item We do not constraint the encoder to place the attribute features in predefined latent variables but we let the triplet encoder, classifier and decoder be free to organize information in the most suitable way.
\end{itemize}

Another line of works aims at combining GANs and VAE to improve image quality or shape the latent space to assume a desired distribution; Adversarial AEs \cite{makhzani2015adversarial}, Wasserstein AEs \cite{tolstikhin2018wasserstein}, VAE-GAN \cite{larsen2016autoencoding} and others \cite{donahue2016adversarial,dumoulin2016adversarially,srivastava2017veegan,ulyanov2018takes} follow such approach. Recently Introspective VAEs (IntroVAEs) \cite{huang2018introvae} have been proposed to improve image quality of VAEs, they exploit the regularization term of the ELBO to set up and adversarial game between the encoder and the decoder achieving image quality comparable to GANs. Our work exploit a similar reasoning to set up a different adversarial game the exploits the linear classifier available in the VAC framework to build our IntroVAC that is able to provide realistic images. 
 
\section{Background}
\subsection{Variational Autoencoders}
\begin{figure}[hbt]
    \centering
    \includegraphics[width=0.4\columnwidth]{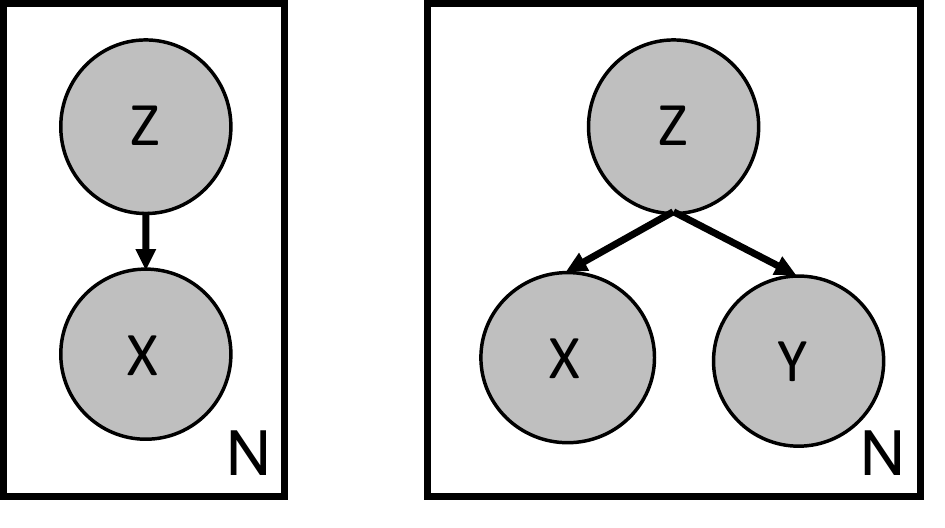}
    \caption{Bayesian network of a VAE (left) and the VAC employed in this work (right).}
    \label{fig:bnet}
\end{figure}

VAEs \cite{higgins2016beta,kingma2013auto} are Deep Generative Models that exploit powerful DNNs to build a complex parametrized model of the data distribution. 
More in details, given a dataset $\mathcal{D}=  \{ \vx_i\}_{i=1}^n$ with data sampled from a unknown probability distribution $ \vx_i \sim p(\vx)$, we aim at finding a parametric model $p_{\vth}(\vx)$ of $p(\vx)$ that maximizes the likelihood of the available data. \\ Introducing a vector of latent variables $\vz \in \mathbb{R}^d$, following the Bayesian network in \cref{fig:bnet} (left), we assume the data to be generated by first sampling  $\vz$ from a known prior $p(\vz) \sim \normst$ and then by sampling $\vx$ from an unknown conditional distribution $p(\vx|\vz)$. Being the latent vectors not available in the dataset, we would like to learn $p_{\vth}(\vx|\vz)$ by maximizing the \textit{marginal likelihood} of the model $p_{\vth}(\vx) = \int p_{\vth}(\vx|\vz)p(\vz)$. However, such integral is intractable for complex $p_{\vth}(\vx|\vz)$ and a Monte Carlo approximation would be computationally too expensive for the dataset sizes typically employed in deep learning settings. Hence, we introduce an approximate posterior $q_{\vphi}(\vz | \vx)$ and maximize the so-called ELBO with respect to the parameters $\vth, \vphi$:
\begin{equation}\label{eq:01}
    \cl_{\vth, \vphi} (\vx) = \bbe_{\qqzx} \left[ \log \left(\ppxz\right) \right] - \kl{\qqzx || \pz};
\end{equation}
in Eq. \ref{eq:01} the first term forces the model to reconstruct the training data while the second term is the Kullback–Leibler divergence between the approximate posterior and the prior, and acts as a regularizer. 
 The learned conditional distributions are parametrized by DNNs, in particular $\ppxz$ is a multivariate Gaussian with mean output of a DNN, $\ppxz \sim \norm{\vmu_{\vth}(\vz), \cdot}$, and acts as a decoder mapping latent variables to the input space (images in this work) while $\qqzx$ is a multivariate Gaussian with both mean and covariance output of a DNN, $\qqzx \sim \norm{\vmu_{\vphi}(\vx), \vsig_{\vphi}(\vx)}$, and acts as an encoder that maps data from the input space to a latent representation. Such parametrization of the posterior allows the regularization term to be computed in closed form while the reconstruction term requires the reparametrization trick to be optimized (see \cite{kingma2013auto}). 
 
 VAEs are very appealing models because they provide a low-dimensional representation of the input data that may be exploited to interpret the model behavior. Of particular interest, is finding representations that reflect human interpretable concepts.
 On the other hand VAEs provide poor quality of the generated/reconstructed images that are typically blurred and lack details.
 
 \subsection{Generative Adversarial Networks}
GANs \cite{goodfellow2014generative} aim at learning a deterministic function that maps the prior distribution $\pz$  over the latent space directly to the data distribution. This is done by solving a two players game where a  network $D_{\vth_{D}}(\vx)$ is trained to detect real samples (images coming from the data distribution) from fake samples generated by a generator network $G_{\vth_{G}}(z)$. The generator tries instead to fool the discriminator in order to generate samples that look real. In particular, the following optimization problem is solved:
 \begin{equation} \label{eq:gan}
     \vth_G, \vth_D =  \argmin_{\vth_G}\argmax_{\vth_D}\bbe_{\pz} \left[ \log(1-D(G(\vz)) \right] +  \bbe_{\px} \left[ \log D(\vx) \right].
 \end{equation}
 
After successful training procedures, the generator is able to provide good quality samples from the data distribution while the discriminator is not able to distinguish between real and generated samples. Contrary to VAEs, images generated from GANs are very realistic and detailed. However GANs do not provide an explicit model of the distribution, hence it is not possible to evaluate the data likelihood; moreover, GANs also lack an encoder network to provide a low dimensional representation of the input. 
\section{Proposed approach}
\begin{figure}
    \centering
    \includegraphics[width=1.0\textwidth]{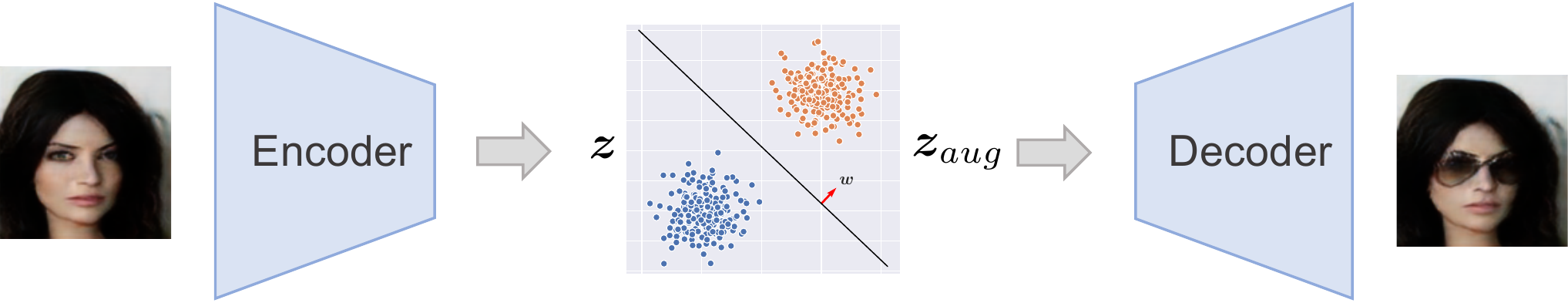}
    \caption{Manipulation process: the encoder provides the latent representation $\vz$ of the image, we move perpendicular to the separating hyperplane in the latent space and then decode the augmented representation $\vz_{aug}$ to get the augmented image.}
    \label{fig:scheme_manipulation}
\end{figure}
 The goal of representation learning is typically to find a set of disentangled latent variables that correspond to specific characteristics of the image such as color, shape etc. In this work, we aim instead at automatically learning directions in the latent space that, when followed, leave the reconstructed image unchanged except for those  specific attributes of the image that characterize a binary class label known for the training set (we assume to have a single binary label to simplify the explanation, however a more general treatise with multiple classes can used) . Formally, we define the dataset as $\mathcal{D}=  \{ \vx_i, y_i\}_{i=1}^n$ with $\vx_i, y_i$ sampled from an unknown joint distribution.
 
 First, we need to extend VAEs to include an additional random variable $y$ representing the class label. Following the Bayesian network in \cref{fig:bnet} (right),  we assume (similarly to VAEs) that the generating process take places by sampling a latent vector  $\vz$ from a known prior $p(\vz) \sim \normst$ and then sampling $\vx$ and $y$ from an unknown conditional distribution $p(\vx, y|\vz)$ that factorizes as $\pxz \pyz$ assuming $\vx$ and $y$  to be  conditional independent given the latent vector; this is reasonable since in order to generate an image $\vx$, the information about its class must be available in $\vz$. Introducing the approximate posterior $\qzxy$ the resulting ELBO is:
 \begin{equation} \label{eq:vacelbo}
     \cl(\vx, y) = \bbe_{\qzxy} \left[ \pxz \pyz \right] - \kl{\qzxy || \pz}.
 \end{equation}
Assuming that all the information about $\vz$ is contained in $\vx$, i.e $\qzxy = \qzx$, the ELBO simplifies to:
  \begin{equation} \label{eq:vacelbos}
  \begin{split}
      \cl(\vx, y) = &\bbe_{\qzx} \left[ \log \right(\pxz \pyz \left) \right] - \kl{\qzx|| \pz} \\
      = &\bbe_{\qzx} \left[ \log \right(\pxz \left) \right] + \bbe_{\qzx} \left[ \log \right(\pyz \left) \right] \\ & - \kl{\qzx|| \pz}
  \end{split}
 \end{equation}
 \cref{eq:vacelbos} is equivalent to VAEs with an additional classification term that encourages the model to assign the correct label to each input sample. More in details, we employ the parametrization $\ppxz \sim \norm{\vmu_{\vth}(\vz), \cdot}$, with $\vmu_{\vth}(\vz),$ output of an decoder network $G_{\vth}(\vz)$, $\qqzx \sim \norm{\vmu_{\vphi}(\vx), \vsig_{\vphi}(\vx)}$ with $\vmu_{\vphi}(\vx), \vsig_{\vphi}(\vx)$ output of an encoder network $E_{\vphi}(\vx)$ while we approximate $\pyz$ with a linear logistic classifier $C(\vz)$ with parameters $\vw$ meaning that $\ppyz = C(\vz)^{\mathbbm{1}[y=1]} + (1-C(\vz))^{\mathbbm{1}[y=0]}$ where $\mathbbm{1}[\cdot]$ is the indicator function. The resulting optimization problem is:
 \begin{equation}\label{eq:param}
 \begin{split}
      \vth, \vphi, \vw = & \argmin_{\vth, \vphi, \vw} \sum_{\vx, y \in \cd}  -\cl_{\vth, \vphi, \vw}(\vx, y) \\
       = & \argmin_{\vth, \vphi, \vw} \sum_{\vx, y \in \cd} \left[ - \bbe_{\qqzx} \left[ \log \right(\ppxz \ppyz \left) \right]  \right. \\  & \left. +  \kl{\qqzx|| \pz} \right] \\
       = & \argmin_{\vth, \vphi, \vw}  \sum_{\vx, y \in \cd} L_{AE} (\vx, \vx_r) + L_{CL}(y, \hat{y}) + L_{reg}(\vz_r),
 \end{split}
 \end{equation}
  \begin{wrapfigure}{R}{0.5\textwidth}
     \centering
     \includegraphics[width=0.48\textwidth]{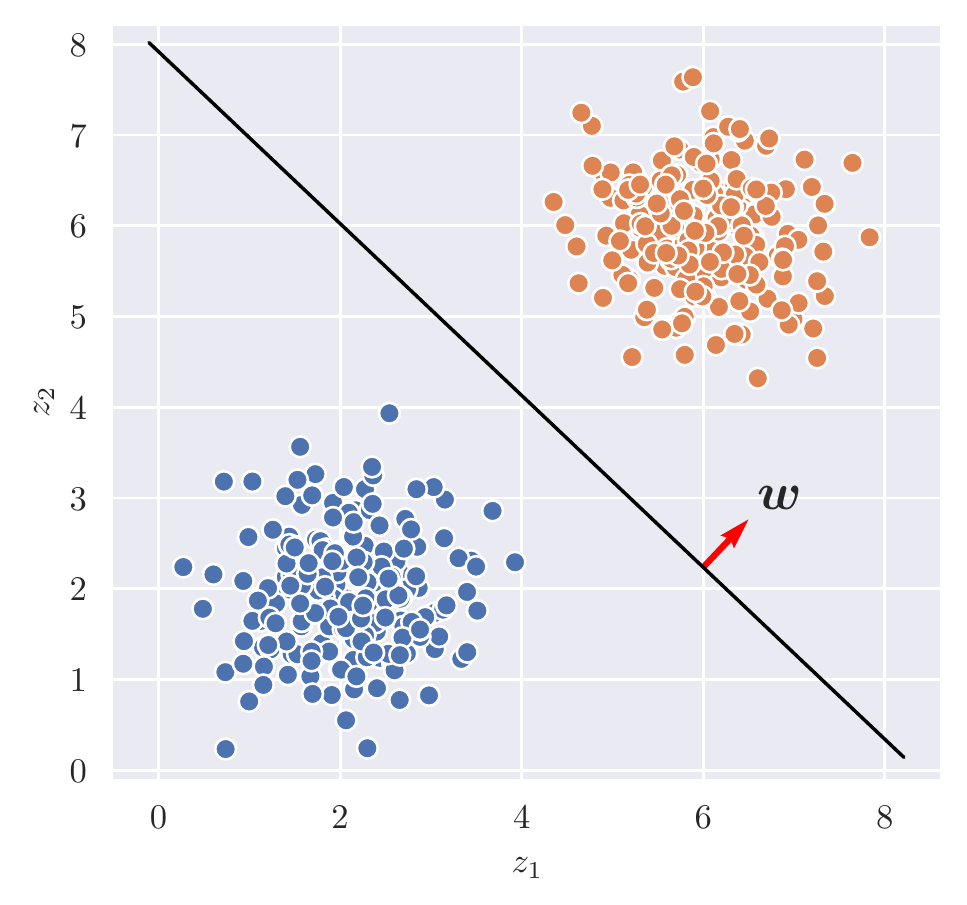}
     \caption{Simple example with a latent space of dimension 2. The parameters vector $\vw$ of the linear classifier points in the direction perpendicular to the separating hyper-plane. Following such direction allows to move from one class to the other.}
     \vspace{-10pt}
     \label{fig:direction}
 \end{wrapfigure}
 
where $L_{AE}$ is the Mean Squared Error (MSE) between $\vx$ and its reconstruction $\vx_r$, $L_{CL}$ is the Binary Cross Entropy (BCE) between the real ($y$) and predicted ($\hat{y}$) label and $L_{reg}(\vz_r)$ is the KL divergence between two Gaussian random variables with diagonal covariance matrix.
 The optimization problem described in Eq. \ref{eq:param} can be solved with standard first-order optimization algorithms used to train DNNs such as Adam. We call this model Variational Classifier (VAC). Note that from now on we won't write explicitly the dependence on the parameters to simplify notation when there is no ambiguity.
 
 The linear classifier on top of the features plays a fundamental role in our approach. The parameters vector $\vw$ immediately provides the direction in the latent space that is related to class attributes (\cref{fig:scheme_manipulation}), being it perpendicular to the separating hyper-plane that distinguishes the different classes (\cref{fig:direction}); while this does not guarantee that other characteristics of the image won't be changed when moving in such direction, we will show empirically that our model actually preserves well the features of the image not related to the class. This is reasonable because during training the linear classifier will try to push the representations to change in a direction parallel to $\vw$ with orientation positive or negative depending on the correct label while the decoder will try to always reconstruct the same image hence it tends to become invariant to changes in such direction that do not affect the label. 
 \subsection{Improving image quality with introspection}
 The reconstruction loss used in VAC still suffers from the same problems of VAEs providing blurry images that lack details and look not realistic. We can however exploit the availability of the classifier to set up an adversarial optimization strategy similar to GANs but without the need of an additional discriminator network. We proceed as follows: we equip the classifier with a new class whose role is to determine whether an image is real or fake, note that this amounts to add an additional logit to $C$ with an increase in number of parameters equal to the size of the latent space (plus one if we add a bias) which is insignificant compared to the total size of the model. In addition to the loss in \cref{eq:vacelbos} we then add two adversarial losses: for reconstructed images, the couple encoder/classifier is trained to detect them as fake, while the decoder tries to fool encoder and classifier to predict the correct class of the original image that has been reconstructed. For generated images instead, the couple encoder/classifier acts in the same way while the decoder simply try to maximize their loss. Of course in this case a label is not available hence the goal of the decoder is only to make the image look real. More in details for reconstructed images given $\vz_r$ the latent variable of the associated real image we have:
 \begin{equation}
 \begin{split}
     \vx_{rr} = & \, G(\text{dt}(\vz_r)) \\
      L^r_{E,C}(\vx_{rr}) = & -\log\left[C^{fake}(E(\text{dt}(\vx_{rr})) \right] \\
      L^r_G(\vx_{rr}, y) = & -y\log\left[C^{class}(E(\vx_{rr})) \right] - (1-y)log\left[1-C^{class}(E(\vx_{rr})) \right]  \\
     & - \log\left[1-C^{fake}(E(\vx_{rr})) \right],
 \end{split}
 \end{equation}
 where we denoted as $C^{fake}$ the logit related to the fake/real class and $C^{class}$ the logit related to the image class, while "dt" stands for "detach" and means that gradients are stopped hence $L_{E,C}$ is not backpropagated through the decoder.
 
 For generated images, let $\vz_g$ be a latent vector sampled from the prior we have:
  \begin{equation}
 \begin{split}
     \vx_{g} =& \, G(\vz_g) \\
      L^g_{E,C}(\vx_g) =& -\log\left[C^{fake}(E(\text{dt}(\vx_{g})) \right] \\
      L^g_G(\vx_g) =& - \log\left[1-C^{fake}(E(\vx_{g})) \right].
 \end{split}
 \end{equation}
 This reasoning is similar to IntroVAEs \cite{huang2018introvae} but, while IntroVAEs exploit the divergence term to build the adversarial loss, we exploit the classifier already available in our model that not only provides real looking images but also encourages the decoder to reconstruct images with the correct class attributes. The resulting training procedure is reported in \cref{alg:training}. We employ a set of weights $\beta_{AE}, \beta_{CL}, \beta_{reg}, \beta_{G}, \beta_{E,C}$ to balance the different losses.

 \begin{algorithm}[!h] 
\caption{IntroVAC Training Procedure - The capital bold notation indicates a tensor containing a batch of data, all the losses are averaged over the batch.}\label{alg:training}
\begin{algorithmic}[1] 
\Input $\mathcal{D}= \{ \vx_i, y_i\}_{i=1}^n, $
\While{not converged}
        \State Sample a mini-batch from the training set $\mx, \my $
		\State Sample a set of latent variables from the prior $\mz_g \sim p(\vz)$
		\State $\mx_g \gets G(\mz_g)$
		\State $\mz_r \gets E(\mx)$
		\State $\hat{\my} \gets C(\mz_r)$
		\State $\mx_r \gets G(\mz_r)$
		\State $\mx_{rr} \gets G(\text{dt}(\mz_r))$
		\State $L \gets \beta_{AE}L_{AE}(\mx, \mx_r) + \beta_{CL}L_{CL}(\my, \hat{\my}) + \beta_{reg}L_{reg}(\mz_r) + \beta_{E,C}(L_{E,C}^r(\mx_{rr}) +L^g_{E,C}(\mx_g))$ 
		\State Compute the gradient of $L$ with respect to $\vth, \vphi, \vw$ 
		\State Perform Adam update on $\vphi, \vw$
		\State $L \gets \beta_{G}(L^r_{G}(\mx_{rr}, \my) + L^g_{G}(\mx_g))$ 
		\State Compute the gradient of $L$ with respect to $\vth$
		\State Perform Adam update on $\vth$
		\State Set gradients to zero
\EndWhile
\end{algorithmic}
\end{algorithm}
\section{Results}
\subsection{Experimental Setting}
We run our experiments on the CelebA dataset with resolution $128\times 128$. We create three smaller datasets: (i) one containing images with glasses and without glasses (called \eye ), (ii) a similar one for the attribute facial hair (beard, mustache, goatee) (called \fac) and (iii) one combining the two (called \efac). Hence, in the first two cases we have a single attribute label while for the last one we have two. We train for 150 epochs using Adam optimizer with learning rate 0.0002; we use batch size 64 and we decreased the learning rate at epoch 60, 90 and 120 by a factor of 2; we use $\beta_{AE}=100, \beta_{CL}=10, \beta_{reg}=3, \beta_{G}=5, \beta_{E,C}=0.01$. The architecture employed is similar to \cite{huang2018introvae}: the encoder has 5 residual layers with 32, 64, 128, 256, 512 channels, after each layer downsampling is performed using average pooling reducing the spatial dimension of the input by half. The last layer is linear and  outputs the mean and the diagonal of the covariance matrix; we adopt a latent space of size 256, hence the output of the encoder is of dimension $2\cdot256=512$. The decoder has a similar architecture to the encoder but mirrored and performs upsampling operations after each residual layer to output an RGB image with size corresponding to the input. The classifier is linear and outputs a vector of logits of size equal to the number of classes plus 1 for the adversarial loss.

For the standard VAC we use the same hyperparameters but we train for 600 epochs and don't perform learning rate decay. In both cases we employ the $L_1$ loss for $L_{AE}$ instead of the MSE since it has been shown to provide sharper images.
\subsection{Attribute manipulation}
In order to show that our model is able to learn meaningful directions in the latent space we proceed as follow \cref{fig:scheme_manipulation}: 
\begin{enumerate}
    \item We feed an image $\vx$ from the test set to the encoder network and obtain the mean of the conditional distribution $\qqzx$: $\vz = E(\vx)$.
    \item Starting from the obtained mean, we move in the direction perpendicular to the separating hyper-plane by a distance $\delta$: $\vz_{aug} = \vz \pm \delta \frac{\vw}{||\vw||}$. We use the $+$ sign  to add an attribute and the $-$ sign to remove it. This is equivalent to ascend or descend the gradient to maximize/minimize the logit.
    \item We feed the obtained latent vector to the decoder to obtain the augmented image: $\vx_{aug} = G(\vz_{aug})$.
\end{enumerate}
We use $\delta = 3$ for the \eye{} dataset and $\delta=4$ for the \fac{} dataset while for the \efac{} dataset we use $\delta=4$ if we want ot add a single attribute and $\delta=5$ to add both. One could also use an automated procedure by progressively moving the latent representation until the reconstructed image is confidently classified with the correct label but in our experiments we use a fixed $\delta$ for simplicity.

\begin{figure}
    \centering
    \includegraphics{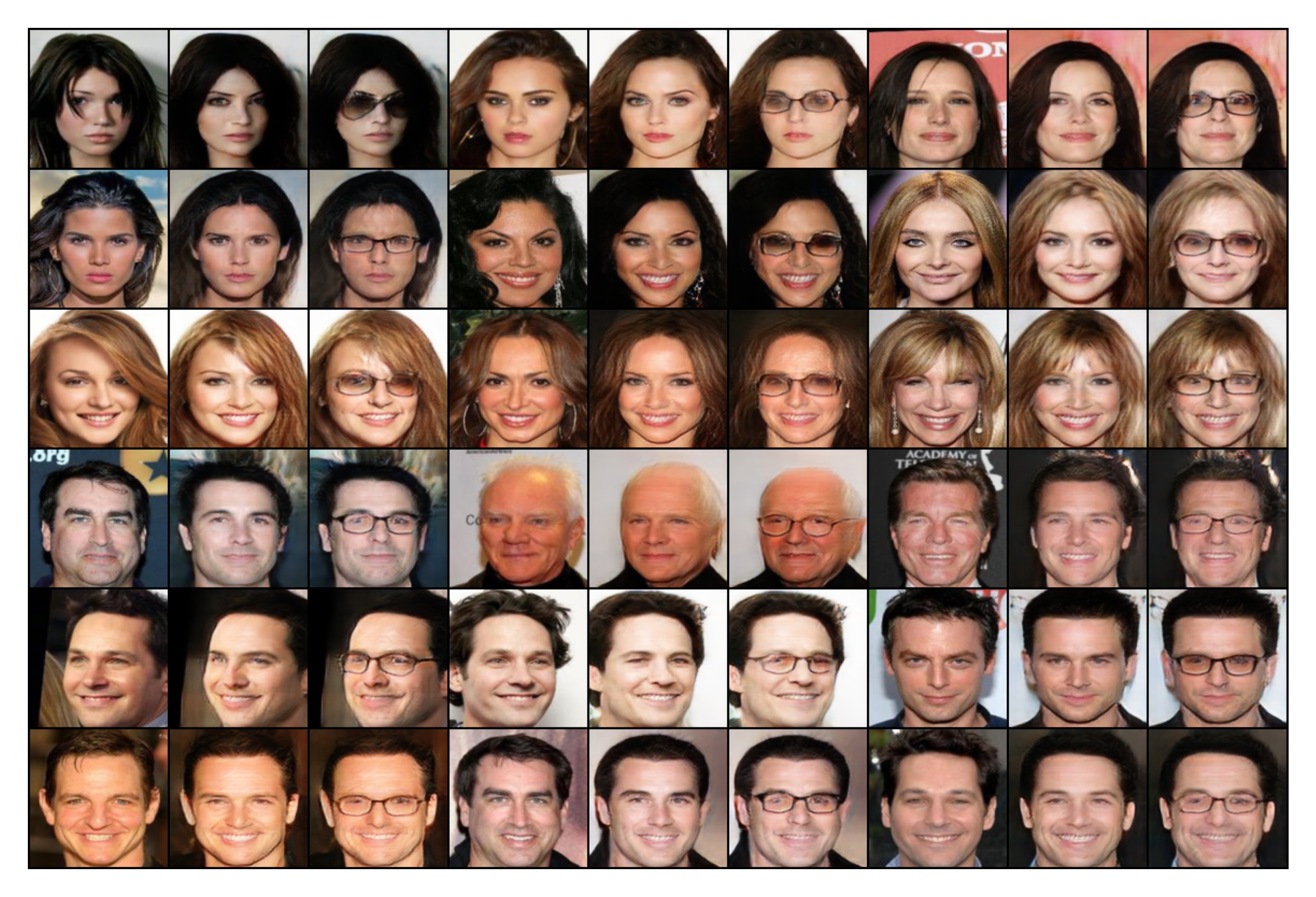}
    \caption{Attribute manipulation results for the \eye{} dataset. For every triplet of images, the one on the left is the original one, the one in the middle is the reconstruction provided by the model and the one on the right is the image reconstructed with the attribute added.}
    \label{fig:glass_transfer}
\end{figure}

We report in \cref{fig:glass_transfer} the results obtained using \eye{} dataset. For every triplet of images, the one on the left is the original one, the one in the middle is the reconstruction provided by the model and the one on the right is the image reconstructed with the attribute added using the procedure described above. Notice how the model is able to provide high detailed reconstructions that looks very realistic. Moreover, the attribute modification does not affect heavily other parts of the image that are not related to the class label.  

\begin{figure}
    \centering
    \includegraphics{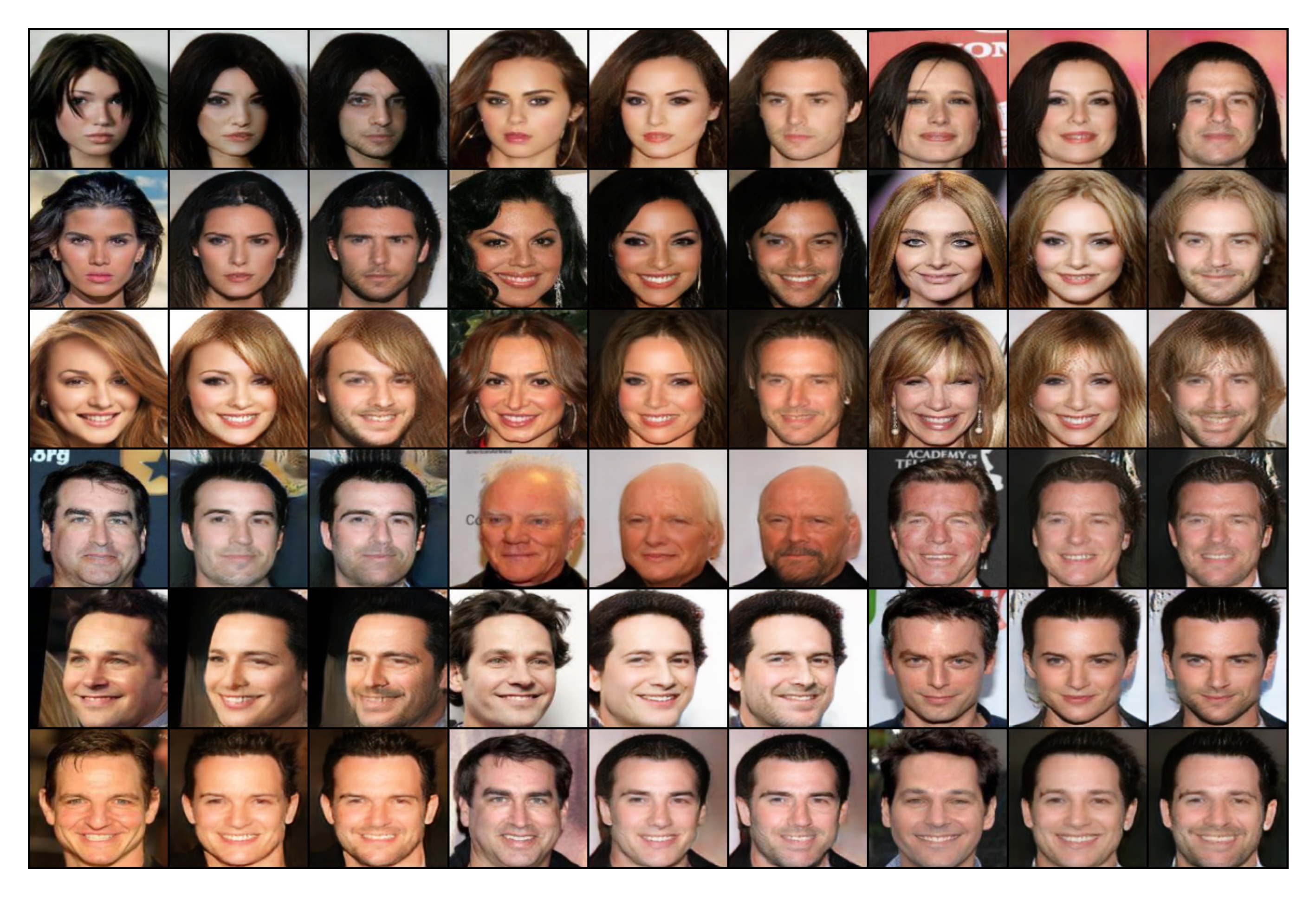}
    \caption{Attribute manipulation results for the \fac{} dataset.}
    \label{fig:hair_transfer}
\end{figure}

Similarly in \cref{fig:hair_transfer} we report results for the \fac{} dataset, also in this case the model is able to clearly add the attribute to the images. Differently from the \eye{} case, there is a very strong correlation between facial hair and gender, hence it is noticeable how women start to assume male traits. This is quite common in such models and it is difficult to remove correlations that are so strong in the dataset.

\begin{figure}
    \centering
    \includegraphics{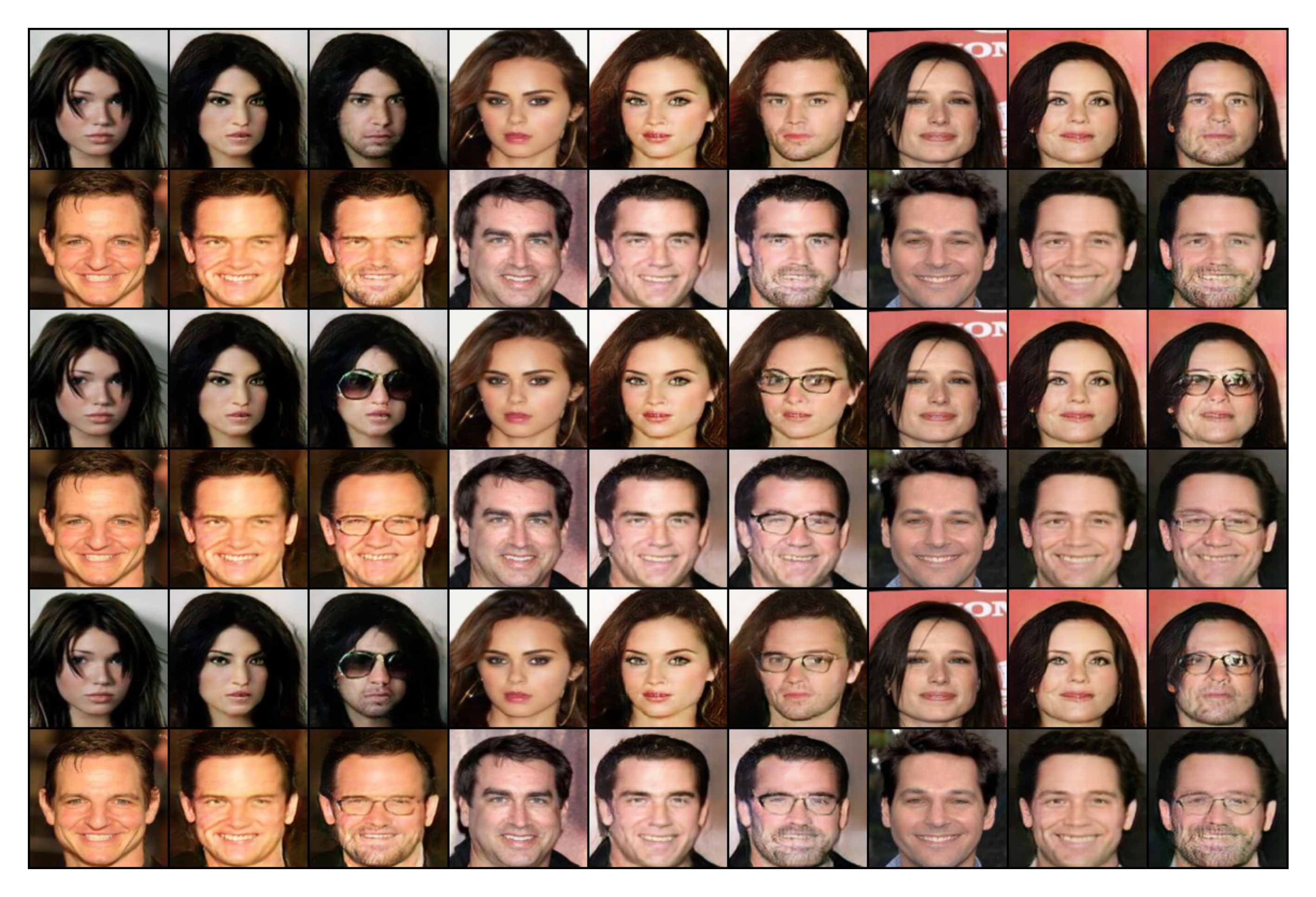}
    \caption{Results for the \efac{} dataset adding the attribute facial hair  (first two rows) eyeglasses  (middle rows) and both (last two rows).}
    \label{fig:eyefacial}
\end{figure}

\cref{fig:eyefacial} shows results with the \efac{} dataset, notice how the model is able to add only the facial hair attribute (first two rows), only the eyeglasses attribute (middle rows) or both the attributes(last two rows).

\begin{figure}
    \centering
    \includegraphics{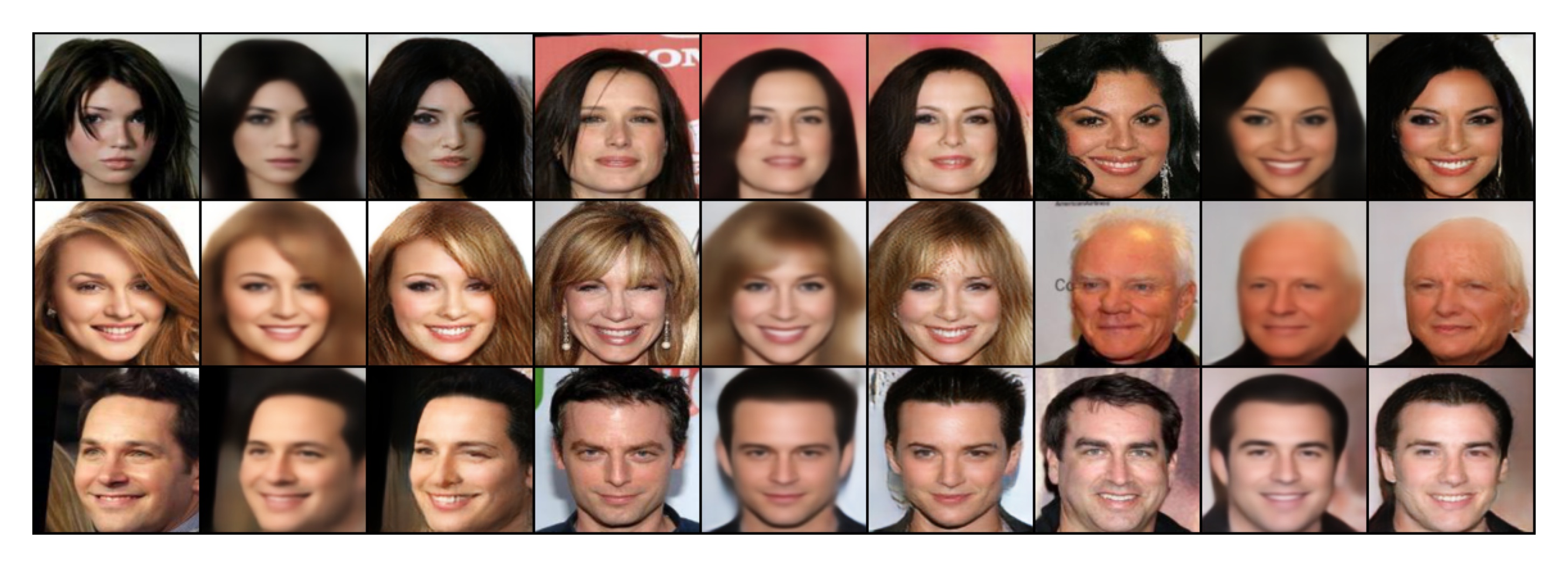}
    \caption{Image quality comparison between our IntroVAC and a standard VAC. For every triplet of images, the one on the left is the original one, the one in the middle is the reconstruction provided by the VAC and the one on the right is the image reconstructed by the IntroVAC. IntroVAC is able to provide much more realistic looking images.}
    \label{fig:quality}
\end{figure}

Eventually, \cref{fig:quality} shows a comparison between a VAC and an IntroVAC, for every triplet of images, the one on the left is the original one, the one in the middle is the reconstruction provided by the VAC and the one on the right is the image reconstructed by the IntroVAC; it is clear how the introspective loss of our model improves quality of reconstructed images that preserve more details and are in general more realistic. To quantitatively prove such statement, we report in \cref{tab:fid} the Fréchet Inception Distance (FID) \cite{heusel2017gans} between the real images and the reconstructed ones for the VAC and IntroVAC models. It is clear how the IntroVAC outperforms VAC.

\begin{table}[]
    \centering
    \begin{tabular}{c|c}
        VAC & IntroVAC \\ \hline
         63.9 & 25.5 
    \end{tabular}
    \caption{FID for the standard VAC and IntroVAC. Lower is better.}
    \label{tab:fid}
\end{table}

\subsection{Conditional image generation with Langevin Monte Carlo}
The linear classifier of VACs provides a parametric model for the conditional distribution of the labels given the latent variables $\ppyz$. We would like to sample images conditioned on a particular class by first sampling latent variables from the posterior distribution $\pzy$ and then feeding them to the decoder. Since we don't have a model for $\pzy$ we resort to Langevin Monte Carlo \cite{hammersley2013monte} methods.

Given an energy function $U(\vz)$ with $\vz \in \bbr^d$ and starting from a point $\vz_0$ random, the iteration:
\begin{equation}\label{eq:langevin}
    \vz_{k+1} = \vz_k - \alpha \nabla U(\vz_k) + \sqrt{2\alpha} \bm{\epsilon}_k
\end{equation}
with $\bm{\epsilon}_k \sim \normst$ a sequence of i.i.d. isotropic Gaussian vectors, converges to the Gibbs distribution $\frac{e^{-U(\vz)}}{\int_{\bbr^m} e^{-U(\vz)}dx}$. Hence, to sample from $\ppzy \propto \ppyz\pz$ we start from a random point sampled from the prior $\vz_0 \sim \pz$ and let the iteration in \cref{eq:langevin} evolve with $U(\vz)=-\log(\ppyz\pz)$ until it converges to the stationary distribution and then collect the samples that are now coming from the correct posterior. Notice that the prior in this case is a quite good starting point for the iteration, hence convergence can be achieved quite rapidly. In practice, we noticed that we achieved better results by starting from a set of different initial conditions and run the iteration for a fixed number of steps. This is equivalent since the stationary distribution is the same. To obtain better samples, we reject the ones that are not classified with the correct attribute.

\begin{figure}
    \centering
    \includegraphics{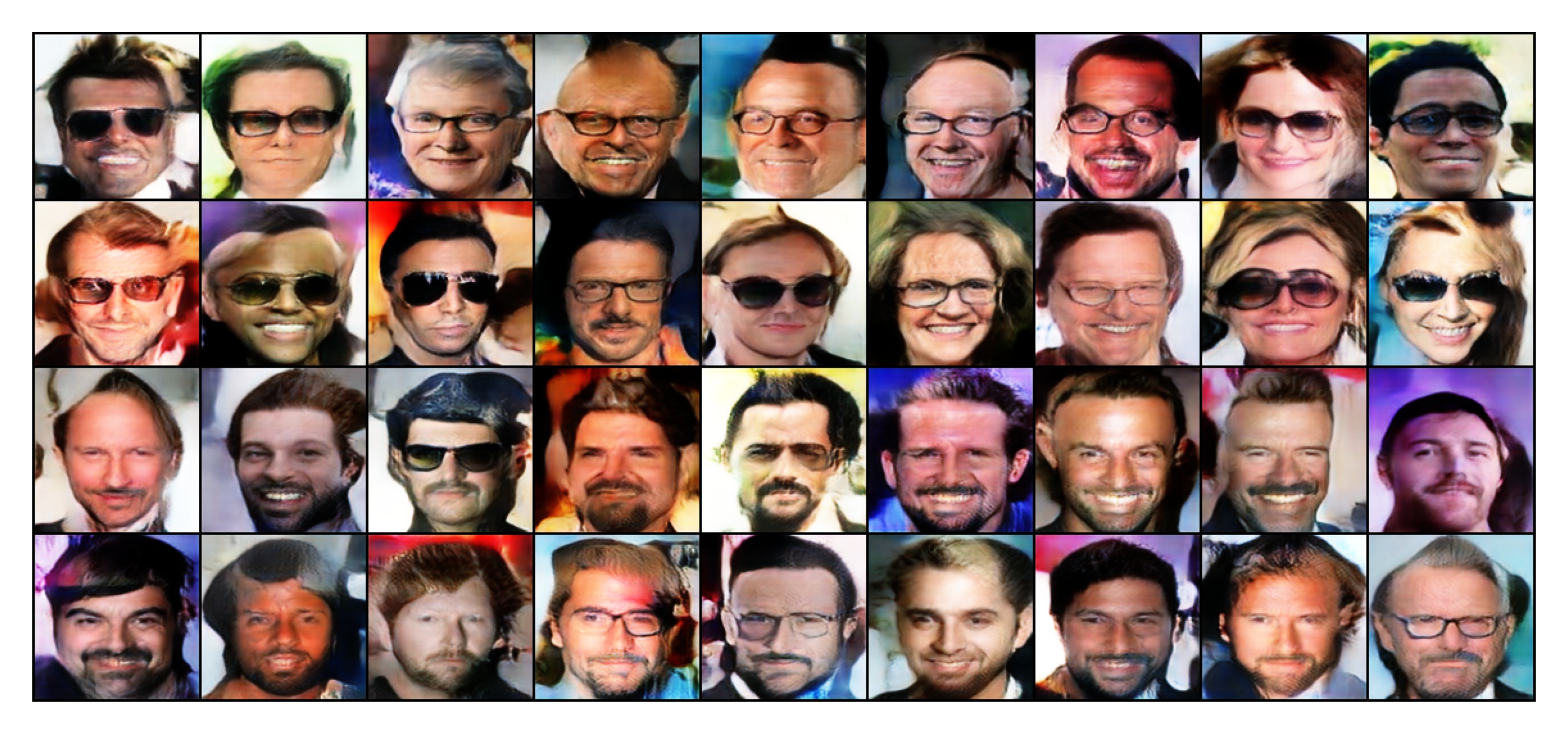}
    \caption{Images generated with Langevin Monte Carlo sampling conditioned on the class eyeglasses (top two rows) and on the class facial hairs (last two rows). }
    \label{fig:condgen}
\end{figure}

In \cref{fig:condgen} we report some samples obtained with this method. The first two rows are conditioned on the attribute glasses and use a model trained on the \eye{} dataset  while the last two are conditioned on the attribute facial hair and use a model trained on the \fac{} dataset. In both cases the attribute is clearly visible. We adopted $\alpha = 0.0002$ and 5000 steps.
\section{Conclusions}
In this paper we introduced IntroVAC, a generative model based on VAEs which, by exploiting information from a class label associated to each training sample, is able to learn meaningful directions in the latent space correlated with class attributes. 
This is achieved by using a simple linear classifier on top of the encoder network whose parameter vector is perpendicular to the separating hyper-plane by construction. Hence, by moving latent representations in a direction parallel to this plane, we can insert or remove class-related attributes on the reconstructed images. 
Moreover, IntroVAC exploits the presence of the linear classifier to perform an adversarial training strategy that is able to improve image quality. We tested our model on three datasets derived from CelebA and showed its ability to perform fine grained attribute manipulation.


\bibliographystyle{plain}
\bibliography{main.bib}

\newpage
\section*{Appendix}
\subsection*{Proof of ELBO}
Let $\pzxy$ be the true posterior over latent variables we have:
\begin{equation}
\begin{split}
    & D_{KL}(\qzxy || \pzxy) = \\
    & =\int_{\bbr^d} \log  \left( \frac{\qzxy}{\pzxy} \right)\qzxy d\vz= \\
    & = \int_{\bbr^d} \log  \left( \frac{\qzxy \pxay}{\pzaxay} \right) \qzxy d\vz= \\
    & =  \int_{\bbr^d} \log  \left(\frac{\qzxy}{\pxyz\pz} \right) \qzxy d\vz + \log \left( \pxay \right)= \\
    & =  -\bbe_{\qzxy} \left[ \pxyz \right] + D_{KL}(\qzxy || \pz)+ \log \left( \pxay \right)=
\end{split}
\end{equation}
Hence since $D_{KL}(\qzxy || \pzxy) \geq 0$ we can bound the marginal likelihood from below as:
\begin{equation}
     \log \left( \pxay \right) \geq  \bbe_{\qzxy} \left[ \pxyz \right] - D_{KL}(\qzxy || \pz)
\end{equation}
\subsection*{Image generation}
We report here some examples of images generated by a model trained on the \efac{} dataset sampling latent variables from the prior.
\begin{figure}[!h]
    \centering
    \includegraphics{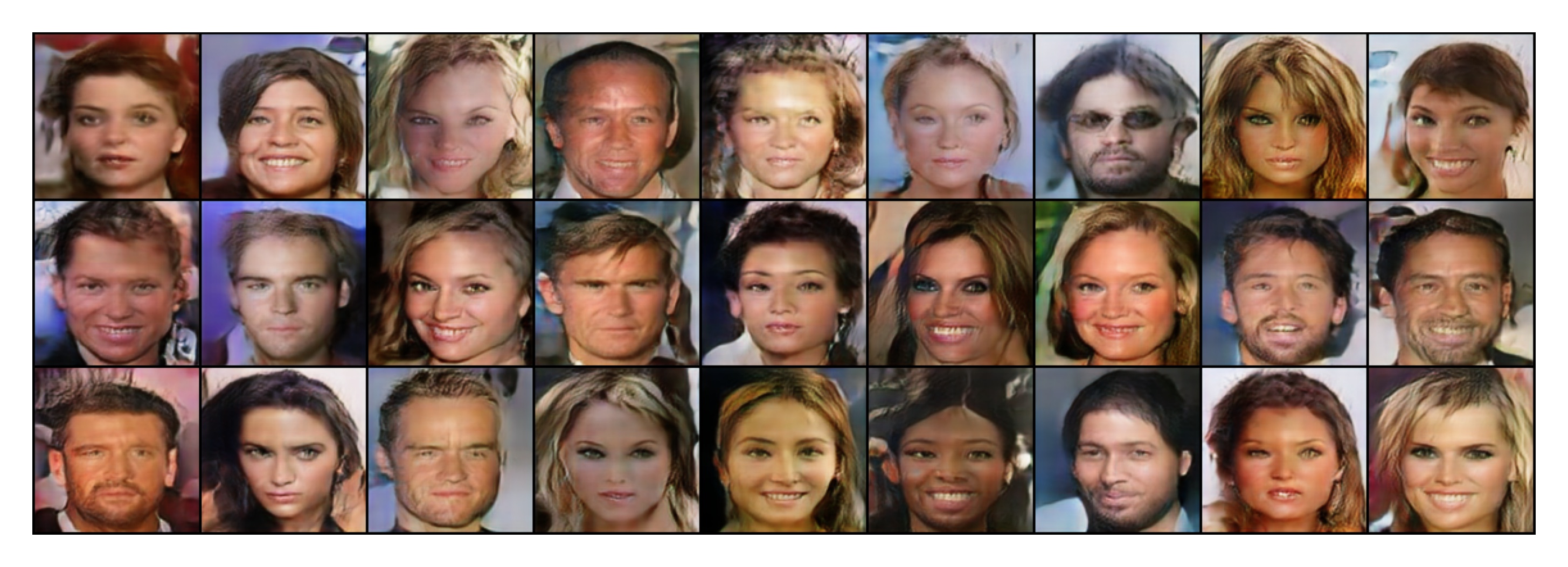}
    \caption{Generated images.}
    \label{fig:gen}
\end{figure}
\newpage
\subsection*{Architecture}
\begin{figure*}[!h]
  	\subfloat[]{%
      		 \includegraphics[width=0.35\textwidth]{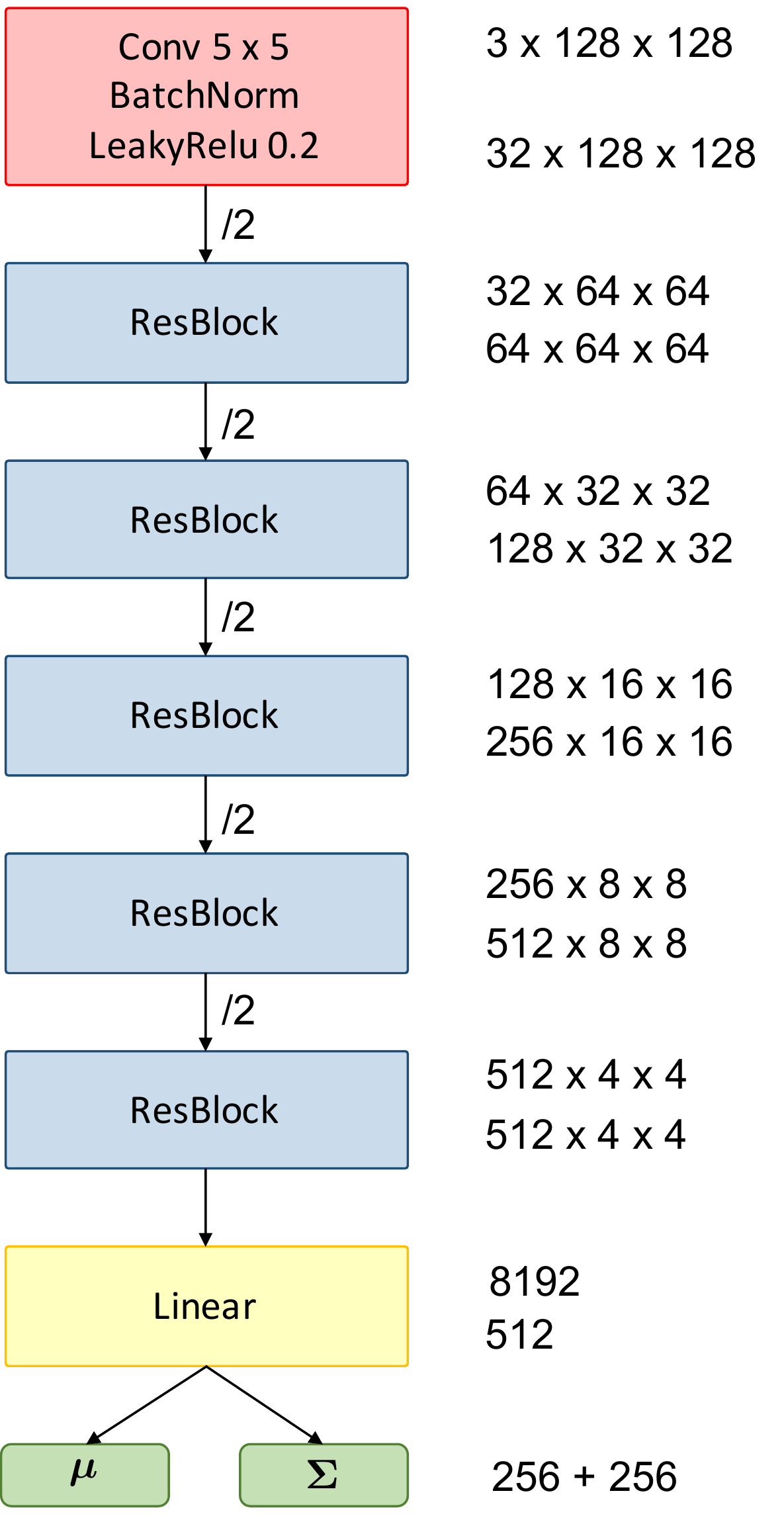}} 
  	\subfloat[]{%
  	 	\includegraphics[width=0.35\textwidth]{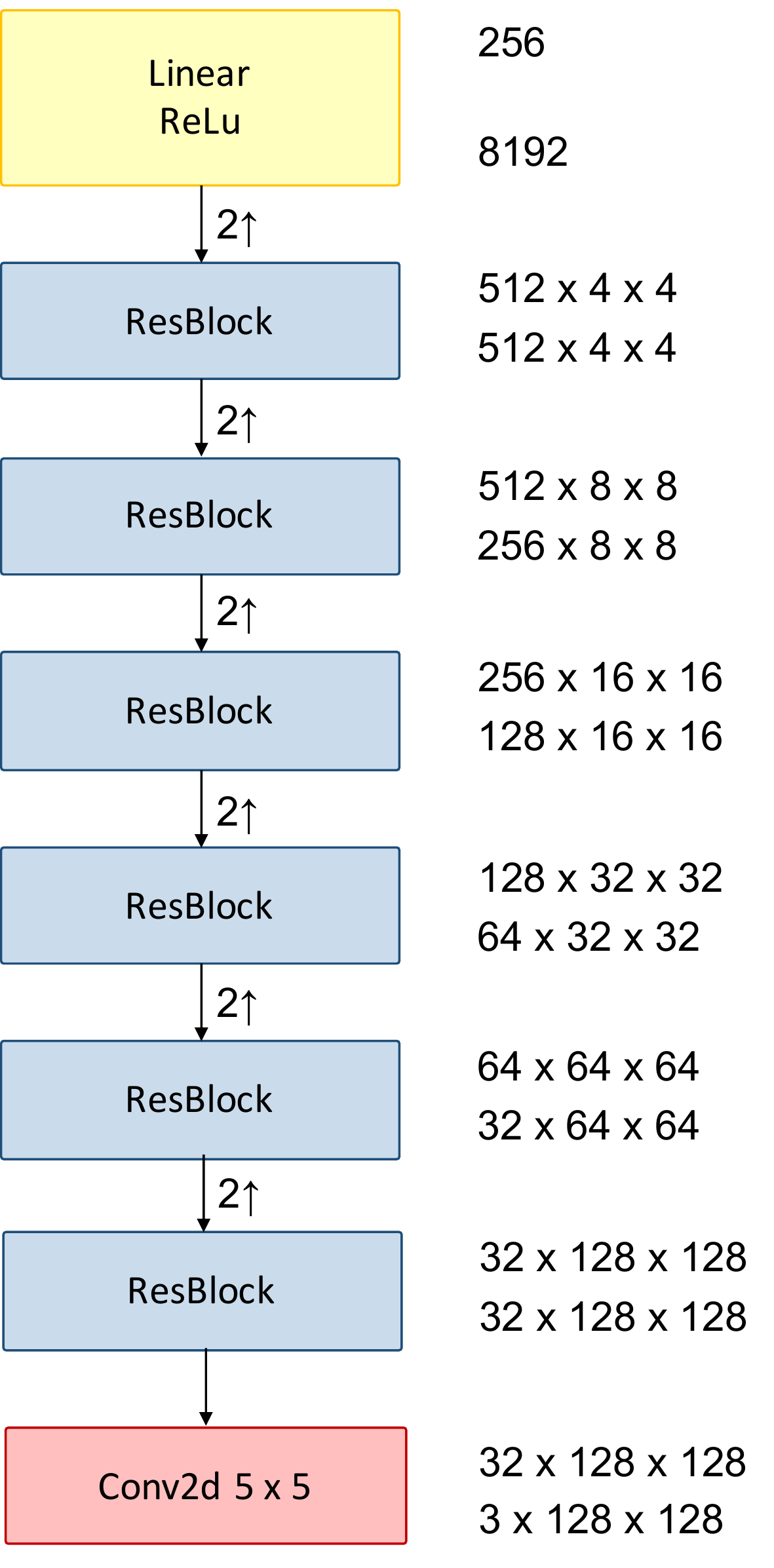}}
  	 	  	\subfloat[]{%
   	\includegraphics[width=0.3\textwidth]{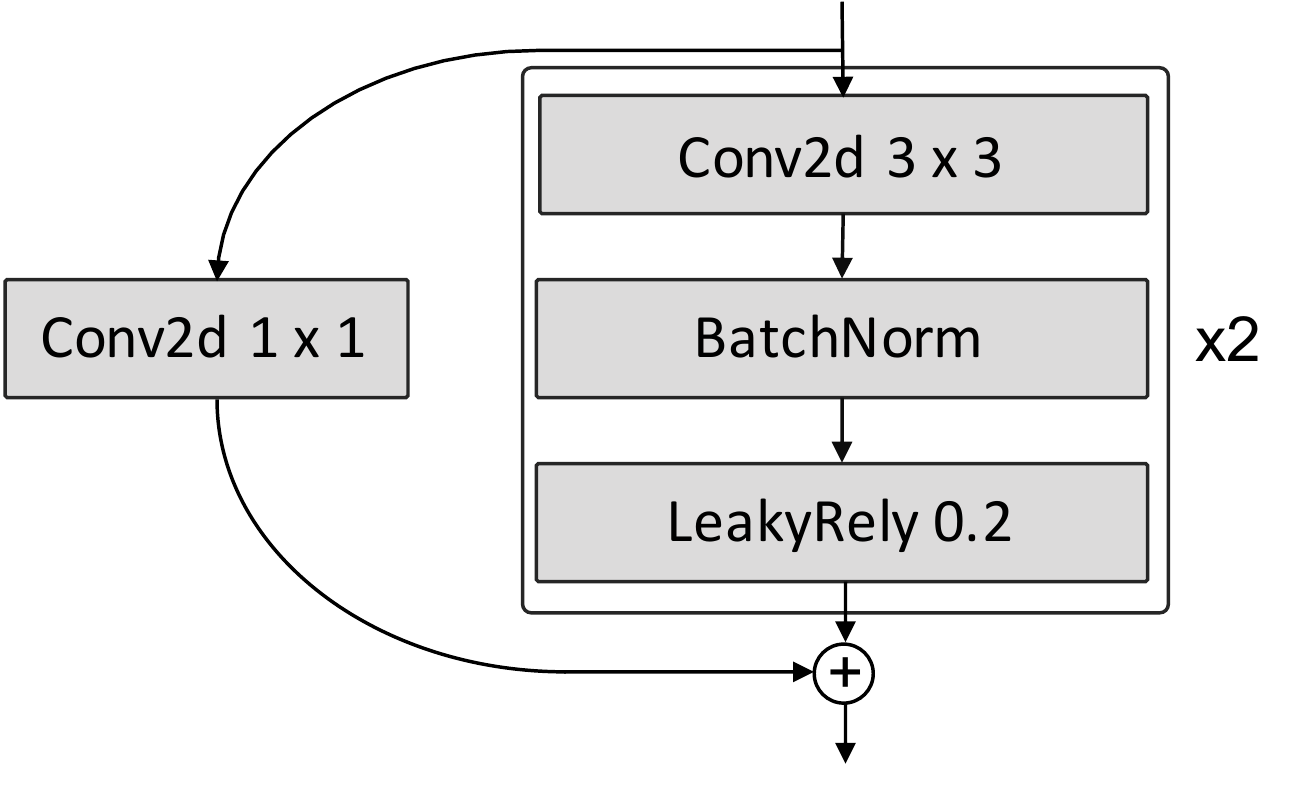}}
 \caption{(a) Structure of the Encoder. (b) Structure of the Decoder. (c) Structure of each residual block (ResBlock) employed in both Encoder and Decoder.}
\label{fig:struct} 
\end{figure*}
\end{document}